\documentclass[10pt,twocolumn,letterpaper]{article}

\usepackage{iccv}
\usepackage{times}
\usepackage{epsfig}
\usepackage{graphicx}
\usepackage{amsmath}
\usepackage{amssymb}
\usepackage{graphics}
\usepackage{subfigure}
\usepackage{amssymb}
\usepackage{amsthm}
\usepackage{algorithm}
\usepackage{algorithmic}
\usepackage{booktabs}
\usepackage{threeparttable}
\usepackage{multirow}

\newcommand{\be}{\mathbf{e}}

\newcommand{\x}{\mathbf{x}}

\newcommand{\cB}{\mathcal{B}}
\newcommand{\cC}{\mathcal{C}}

\newcommand{\cP}{\mathcal{P}}

\newcommand{\cT}{\mathcal{T}}

\newcommand{\real}{\mathbb{R}}

\newcommand{\udots}{\mathinner{\mskip1mu\raise1pt\vbox{\kern7pt\hbox{.}}\mskip2mu\raise4pt\hbox{.}\mskip2mu\raise7pt\hbox{.}\mskip1mu}}

\newcommand\aka{\emph{a.k.a.}}

\usepackage[breaklinks=true,bookmarks=false]{hyperref}

\iccvfinalcopy 

\def\iccvPaperID{****} 

\begin{document}

\title{Deep Metric Learning with Angular Loss}

\author{Jian Wang, Feng Zhou, Shilei Wen, Xiao Liu and Yuanqing Lin\\
Baidu Research\\
{\tt\small \{wangjian33,zhoufeng09,wenshilei,liuxiao12,linyuanqing\}@baidu.com}
}

\maketitle

\begin{abstract}

The modern image search system requires semantic understanding of image,
and a key yet under-addressed problem is to learn a good metric for measuring the similarity between images.
While deep metric learning has yielded impressive performance gains by extracting high level abstractions from image data,
a proper objective loss function becomes the central issue to boost the performance.
In this paper, we propose a novel angular loss, which takes angle relationship into account, for learning better similarity metric.
Whereas previous metric learning methods focus on optimizing the similarity (contrastive loss) or relative similarity (triplet loss) of image pairs,
our proposed method aims at constraining the angle at the negative point of triplet triangles.
Several favorable properties are observed when compared with conventional methods.
First, scale invariance is introduced, improving the robustness of objective against feature variance.
Second, a third-order geometric constraint is inherently imposed, capturing additional local structure of triplet triangles than contrastive loss or triplet loss.
Third, better convergence has been demonstrated by experiments on three publicly available datasets.


\end{abstract}

\section{Introduction}

Metric learning for computer vision aims at finding appropriate similarity measurements between pairs of images that preserve desired distance structure.
A good similarity can improve the performance of image search, particularly when the number of categories is very large~\cite{BhatiaJKVJ15} or unknown.
Classical metric learning methods studied the case of finding a better Mahalanobis distance in linear space.
However, linear transformation has a limited number of parameters and cannot model high-order correlations between the original data dimensions.
With the ability of directly learning non-linear feature representation, deep metric learning has achieved promising results on various tasks, such as visual product search~\cite{Bala15,LiSQFCG15,KiapourHLBB15}, face recognition~\cite{ChopraHL05,TaigmanYRW14,SchroffKP15}, feature matching~\cite{ChoyGSC16}, fine-grained image classification~\cite{WangSLRWPCW14,ZhangZLZ16}, zero-shot learning~\cite{FromeCSBDRM13,WestonBU11} and collaborative filtering~\cite{HsiehYCLBE17}.

\begin{figure}
\centering\includegraphics[width=.5\textwidth]{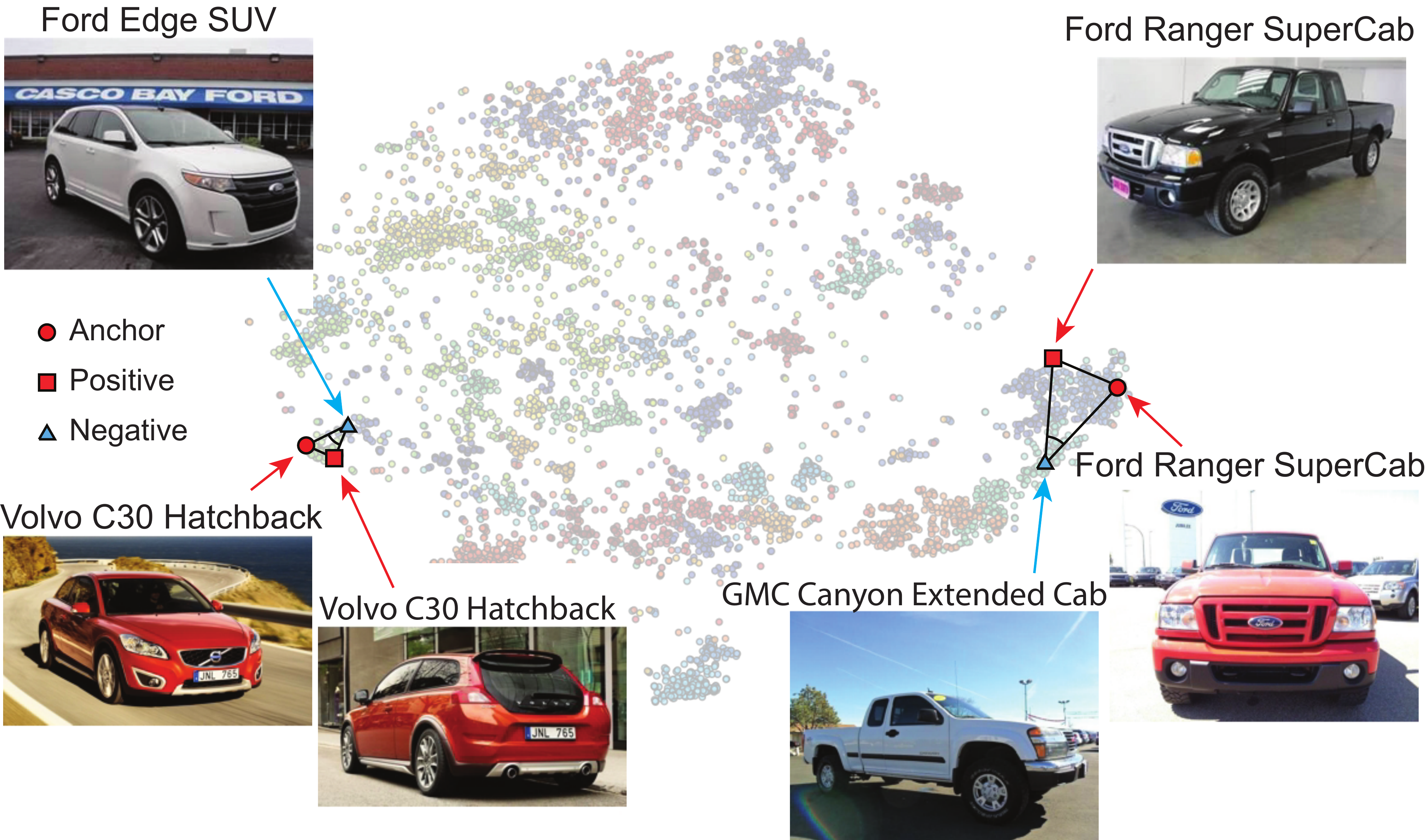}
\caption{Example of feature embedding computed by t-SNE~\cite{Maaten14} for the Stanford car dataset~\cite{KrauseSDF013}, where the images of Ford Ranger SuperCab (right) have a more diverse distribution than Volvo C30 Hatchback (left). Conventional triplet loss has difficulty in dealing with such unbalanced intra-class variation. The proposed angular loss addresses this issue by minimizing the scale-invariant angle at the negative point.} \label{fig:teaser}
\end{figure}

Despite the various forms, the major work of deep metric learning can be categorized as minimizing either the contrastive loss (\aka, Siamese network)~\cite{ChopraHL05} or the triplet loss~\cite{ref8_lx,ref9_lx}.
However, it has been widely noticed that directly optimizing distance-based objectives in deep learning framework is difficult, requiring many practical tricks, such as multi-task learning~\cite{Bala15,ZhangZLZ16} or hard negative mining~\cite{WangSLRWPCW14,CuiZLB16}.
Recent work including the lifted structure~\cite{SongXJS16} and the N-pair loss~\cite{Sohn16} proposed to more effectively mine relations among samples within a mini-batch.
Nevertheless, all of these works rely on certain distance measurement between pairs of similar and dis-similar images.
We hypothesize that the difficulty of training deep metric learning also comes from the limitation by defining the objective only in distance.
First, distance metric is sensitive to scale change.
Traditional triplet loss constrains the distance gap between dis-similar clusters.
However, it is inappropriate to choose the same absolute margin for clusters in different scales of intra-class variation.
For instance, Fig.~\ref{fig:teaser} shows the t-SNE~\cite{Maaten14} feature embedding of Stanford car dataset~\cite{KrauseSDF013}, where the sample distribution of Ford Ranger SuperCabs is much more diverse than Volvo C30 Hatchback.
Second, distance only considers second-order information between samples.
Optimizing distance-based objectives in stochastic training leads to sub-optimal convergence in high-order solution space.

To circumvent these issues, we propose a novel angular loss to augment conventional distance metric learning.
The main idea is to encode the third-order relation inside triplet in terms of the angle at the negative point.
By constraining the upper bound of the angle, our method pushes the negative point away from the center of positive cluster, and drags the positive points closer to each other.
Our idea is analogous to the usage of high-order information for augmenting pair-wise constraints in the domain of graph matching~\cite{DuchenneBKP11} and Markov random fields~\cite{FixGBZ11}.
To the best of our knowledge, this is the first work to explore angular constraints in deep metric learning.
In particular, the proposed angular loss improves traditional distance-based loss in two aspects.
First, compared to distance-based metric, angle is not only rotation-invariant but also scale-invariant by nature.
This renders the objective more robust against the variation of local feature map.
For instance, the two triplets shown in Fig.~\ref{fig:teaser} are quite different in their scales.
It is more reasonable to constrain the angle that is proportional to the relative ratio between Euclidean distances.
Second, angle defines the third-order triangulation among three points.
Given the same triplet, angular loss describes its local structure more precisely than distance-based triplet loss.
Our idea is general and can be potentially combined with existing metric learning frameworks.
The experimental study shows it achieves substantial improvement over state-of-the-arts methods on several benchmark datasets.

\section{Related work}

Metric learning has been a long-standing problem in machine learning and computer vision.
The simplest form of metric learning may be considered as learning the Mahalanobis distance between pairs of points.
It has a deep connection with classical dimension reduction methods such as PCA, LLE and clustering problems but in a discriminative setting.
An exhaustive review of previous work is beyond the scope of this paper.
We refer to the survey of Kulis \etal~\cite{Kulis13} on early works of metric learning.
Here we focus on the two main streams in deep metric learning, contrastive embedding and triplet embedding, and their recent variants used in computer vision.

The seminal work of Siamese network~\cite{BromleyGLSS93} consists of two identical sub-networks that learn contrastive embedding from a pair of samples.
The distance between a positive pair is minimized and small distance between a negative pair is penalized, such that the derived distance metric should be smaller for pairs from the same class, and larger for pairs from different classes.
It was originally designed for signature verification~\cite{BromleyGLSS93}, but gained a lot of attention recently due to its superior performance in face verification~\cite{ChopraHL05,TaigmanYRW14,SunCWT14,YiLL14}.

Despite its great success, contrastive embedding requires that training data contains real-valued precise pair-wise similarities or distances, which is usually not available in practice.
To address this issue, triplet embedding \cite{ref1_lx} is proposed to explore the relative similarity of different pairs and it has been widely used in image retrieval \cite{WangSLRWPCW14, ref9_lx} and face recognition \cite{SchroffKP15}.
A triplet is made up of three samples from two different classes, that jointly constitute a positive pair and a negative pair.
The positive pair distance is encouraged to be smaller than the negative pair distance, and a soft nearest neighbor classification margin is maximized by optimizing a hinge loss.

Compared to softmax loss, it has been shown that Siamese network or triplet loss is much more difficult to train in practice.
To make learning more effective and efficient, hard sample mining which only focuses on a subset of samples that are considered hard is usually employed.
For instance, FaceNet~\cite{SchroffKP15} suggested an online strategy by associating each positive pair in the minibatch with a semi-hard negative example.
Wang \etal~\cite{WangSLRWPCW14} designed a more effective sampling strategy to draw out-class and in-class negative images to avoid overfitting for training triplet loss.
To more effectively bootstrap a large flower dataset, Cui \etal~\cite{CuiZLB16} utilized the hard negative images labeled by humans, which are often neglected in traditional dataset construction.
Huang \etal~\cite{HuangLT16} introduced a position-dependent deep metric unit, which can be used to select hard samples to guide the deep embedding learning in an online and robust manner.
More recently, Yuan \etal~\cite{YuanYZ16} proposed a cascade framework that can mine hard examples with increasing complexities.

Recently, there are also some works on designing new loss functions for deep metric embedding.
A simple yet effective way is to jointly train embedding loss with classification loss.
With additional supervision, the improvement of triplet loss has been evidenced in face verification~\cite{SunCWT14}, fine-grained object recognition~\cite{ZhangZLZ16} and product search problems~\cite{Bala15}.
However, these methods still suffer from the limitation of the conventional sampling that focuses only on the relation within each triplet.
To fix this issue, Song \etal \cite{SongXJS16} proposed the lifted structure to enable updating dense pair combinations in the mini-batch.
Sohn \cite{Sohn16} further extended the triplet loss into N-pair loss, which significantly improves upon the triplet loss by pushing away multiple negative examples jointly at each update.
In addition to these efforts that only explore local relation inside each mini-batch, another direction of work is designed to optimize clustering-like metric that is aware of the global structure of all training data.
Early methods such as neighborhood components analysis (NCA)~\cite{ref10_lx, ref1_lx} can directly optimize leave-one-out nearest-neighbor classification loss.
When applied to mini-batch training, however, NCA is limited as it requires to see the entire training data in each iteration.
Rippel \etal \cite{ref7_lx} improved NCA by maintaining an model of the distributions of the different classes in feature space.
The class distribution overlap is then penalized to achieve discrimination.
More recently, Song \etal~\cite{SongJR016} proposed a new metric learning framework which encourages the network to learn an embedding function that directly optimizes a clustering quality metric.
Nevertheless, all above-mentioned losses are defined in term of distances of points, and very few~\cite{UstinovaL16} has considered other possible forms of loss.
Our work re-defines the core component of metric learning loss using angle instead of distance, and we show it can be easily adapted into existing architectures such as N-pair loss to further improve their performance.

\section{Proposed method}

In this section, we present a novel angular loss to augment conventional deep metric learning.
We first review the conventional triplet loss in its mathematical form.
We then derive the angular loss by constructing a stable triplet triangle.
Finally, we detail the optimization of the angular loss on a mini-batch.

\subsection{Review of triplet loss}

\begin{figure}
\centering\includegraphics[width=.4\textwidth]{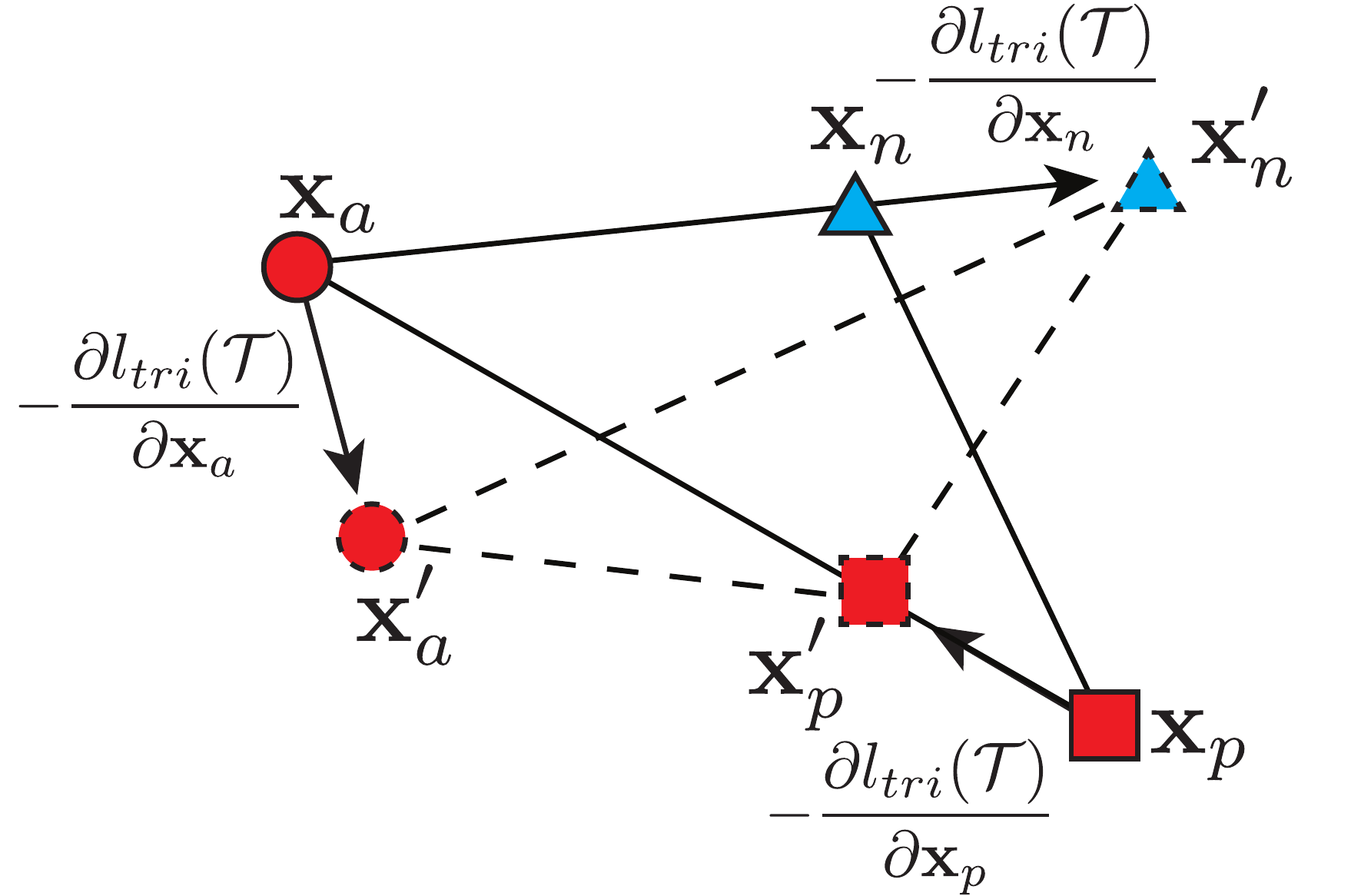}
\caption{Illustration of the triplet loss and its gradient on a synthetic example.} \label{fig:triplet}
\end{figure}

Suppose that we are given a set of training images $\{(\x, y), \cdots\}$ of $K$ classes, where $\x \in \real^D$ denotes the feature embedding of each sample extracted by CNN and $y \in \{ 1,  \cdots, K\}$ its label.
At each training iteration, we sample a mini-batch of triplets, each of which $\cT = (\x_a, \x_p, \x_n)$ consists of an anchor point $\x_a$, associated with a pair of positive $\x_p$ and negative $\x_n$ samples, whose labels satisfy $y_a = y_p \neq y_n$.
The goal of triplet loss is to push away the negative point $\x_n$ from the anchor $\x_a$ by a distance margin $m > 0$ compared to the positive $\x_p$:
\begin{align} \label{eq:triplet_constraint}
  \|\x_a - \x_p \|^{2} + m \leq \| \x_a - \x_n \|^{2}.
\end{align}
For instance, as shown in Fig.~\ref{fig:triplet}, we expect the anchor $\x_a$ to stay closer to the positive $\x_p$ compared to the negative $\x_n$.
To enforce this constraint, a common relaxation of Eq.~\ref{eq:triplet_constraint} is the minimization of the following hinge loss,
\begin{align} \label{eq:triplet_loss}
  l_{tri}(\cT) = \Big[ \|\x_a - \x_p \|^{2} - \| \x_a - \x_n \|^{2} + m \Big]_+,
\end{align}
where the operator $[\cdot]_+ = \max(0, \cdot)$ denotes the hinge function.
It is worth mentioning that the feature map often needs to be normalized to have unit length, \ie, $\| \x \| = 1$, in order to be robust to the variation in image illumination and contrast.

To optimize Eq.~\ref{eq:triplet_loss}, we can calculate its gradient with respect to the three samples of triplet respectively as
\begin{align}
  \frac{\partial l_{tri}(\cT)}{\partial \x_n} &= 2(\x_a - \x_n), \label{eq:triplet_gradient} \\
  \frac{\partial l_{tri}(\cT)}{\partial \x_p} &= 2(\x_p - \x_a), \nonumber \\
  \frac{\partial l_{tri}(\cT)}{\partial \x_a} &= 2(\x_n - \x_p), \nonumber
\end{align}
if the constraint (Eq.~\ref{eq:triplet_constraint}) is violated, or zero otherwise.

It is widely observed that stochastic gradient descent converges poorly on optimizing the triplet loss.
There are a few reasons contributing to this difficulty:
First, it is impractical to enumerate all possible triplets due to the cubic sampling size. Therefore, it calls for an effective sampling strategy to ensure the triplet quality and learning efficiency.
Second, the goal of the objective (Eq.~\ref{eq:triplet_loss}) is to separate clusters by a distance margin $m$.
However, it is inappropriate to apply the single global margin $m$ on the inter-class gap as the intra-class distance can vary dramatically in real-world tasks.
Third, the gradient (Eq.~\ref{eq:triplet_gradient}) derived for each point only takes its pair-wise relation with the second point, but fails to consider the interaction with the third point.
Consider the negative point $\x_n$ in Fig.~\ref{fig:triplet} for an example.
Its gradient $2(\x_a - \x_n)$ may not be optimal without the guarantee of moving away from the class which both the anchor $\x_a$ and positive sample $\x_p$ belong to.

\subsection{Angular loss}

To alleviate the problems elaborated above, a variety of techniques~\cite{Bala15,ZhangZLZ16,WangSLRWPCW14,CuiZLB16,SongXJS16,Sohn16} have been proposed in the last few years.
However, the fundamental component in the loss definition, \ie, the pair-wise distance between points, has rarely been changed.
Instead, this section introduces an angular loss that leads to a novel solution to improve deep metric learning.

\begin{figure}
\centering\includegraphics[width=.5\textwidth]{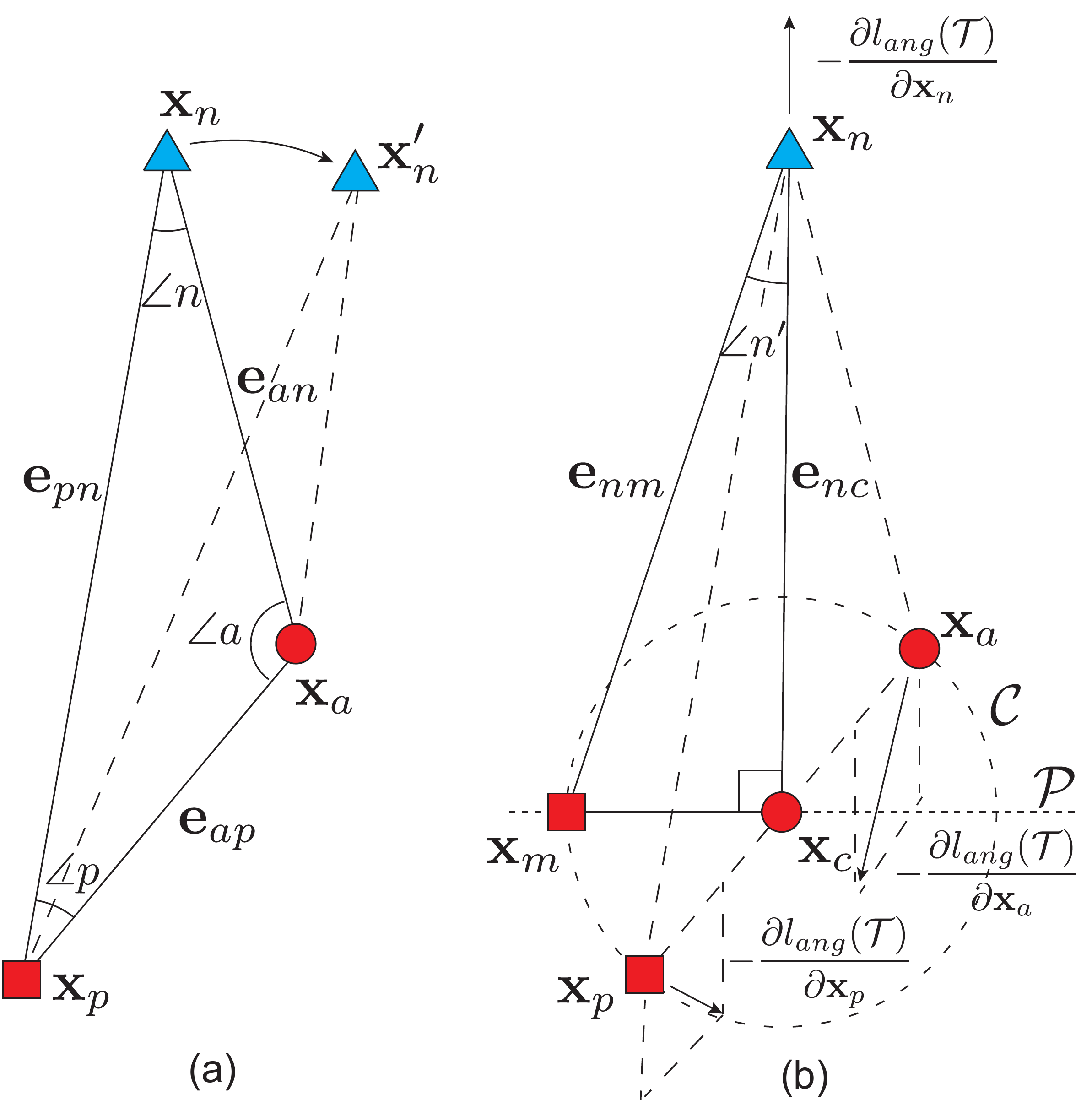}
\caption{Illustration of the angular constraint on a synthetic triplet where $\angle a > 90^{\circ}$.
  (a) Directly minimizing $\angle n$ is unstable as it would drag $\x_n$ closer to $\x_a$. (b) The more stable $\angle n'$ defined by re-constructing the triangle $\triangle_{mcn}$.} \label{fig:toy_example}
\end{figure}

Let's first consider the triplet example shown in Fig.~\ref{fig:toy_example}a, where the triplet $\cT = (\x_a, \x_p, \x_n)$ forms the triangle $\triangle apn$, whose edges are denoted as $\be_{an} = \x_a - \x_n$, $\be_{pn} = \x_p - \x_n$ and $\be_{ap} = \x_a - \x_p$ respectively.
The original triplet constraint (Eq.~\ref{eq:triplet_constraint}) penalizes a longer edge $\be_{an}$ compared to the one $\be_{ap}$ on the bottom.
Because the anchor and positive samples share the same label, we can derive a symmetrical triplet constraint that enforces $\| \be_{ap} \| + m \leq \| \be_{pn} \| $.
According to the cosine rule, it can be proved that the angle $\angle n$ surrounded by the longer edges $\be_{an}$ and $\be_{pn}$ has to be the smallest one, \ie, $\angle n \leq \min(\angle a, \angle p)$.
Furthermore, because $\angle n + \angle a + \angle p = 180^{\circ}$, $\angle n$  has to be less than $60^{\circ}$.
This fact motivates us to constrain the upper bound of $\angle n$ for each triplet triangle,
\begin{align} \label{eq:angular_constraint}
  \angle n \leq \alpha,
\end{align}
where $\alpha > 0$ is a pre-defined parameter.
Intuitively, this constraint selects the triplet that forms a skinny triangle whose shortest edge $\be_{ap}$ connects nodes of the same class.
Compared to the traditional constraint (Eq.~\ref{eq:triplet_constraint}) that is defined on the absolute distance between points, the proposed angular constraint offers three advantages:
1) Angle is a similarity-transform-invariant metric, proportional to the relative comparison of triangle edges.
With a fixed margin $\alpha$, Eq.~\ref{eq:angular_constraint} always holds for any re-scaling of the local feature map.
2) The cosine rule determines the calculation of $\angle n$ involves all the three edges of the triangle.
In contrast, the original triplet only takes two edges into account.
The additional constraint improves the robustness and effectiveness of the optimization.
3) In the original triplet constraint (Eq.~\ref{eq:triplet_constraint}), it is difficult to choose a proper distance margin $m$ without meaningful reference.
By comparison, setting $\alpha$ in the angular constraint is an easier task because it has concrete and interpretable meaning in geometry.

However, a straightforward implementation of Eq.~\ref{eq:angular_constraint} becomes unstable in some special case.
Consider the triangle shown in Fig.~\ref{fig:toy_example}a, where $\angle a > 90^{\circ}$.
By enforcing Eq.~\ref{eq:angular_constraint} to reduce $\angle n$, the negative point $\x_n$ would be potentially dragged towards $\x_n'$, which is closer to the anchor point $\x_a$.
This result contradicts our original goal of enlarging the distance between points of different classes.
To fix this issue, we re-construct the triplet triangle to make Eq.~\ref{eq:angular_constraint} more stable.
Our intuition is to model the relation between the negative $\x_n$ with the local sample distribution defined by the anchor $\x_a$ and the positive $\x_p$, shown in Fig.~\ref{fig:toy_example}b.
A natural approximation to this distribution is the circumcircle $\cC$ passing through $\x_a$ and $\x_p$, centered at the middle $\x_c = (\x_a + \x_p) / 2$.
We then introduce a hyper-plane $\cP$, which is perpendicular to the edge $\be_{nc} = \x_n - \x_c$ at $\x_c$.
The hyper-plane $\cP$ intersects the circumcircle $\cC$ at two nodes, one of which is denoted as $\x_m$.
Based on these auxiliary structures, we define the new triangle $\triangle_{mcn}$ by shifting the anchor $\x_a$ and positive $\x_p$ to $\x_c$ and $\x_m$ respectively.
Given the new triangle, we re-formulate Eq.~\ref{eq:angular_constraint} to constrain the angle $\angle n'$ closed by the edge of $\be_{nc}$ and $\be_{nm}$ to be less than a pre-define upper bound $\alpha$, \ie,
\begin{align} \label{eq:angular_constraint2}
   \tan \angle n' = \frac{\|\x_m - \x_c \|}{\|\x_n - \x_c\|} = \frac{\| \x_a - \x_p \|}{ 2 \|\x_n - \x_c\|} \leq \tan \alpha,
\end{align}
where $\|\x_m - \x_c \|$ is the radius of the circumcircle $\cC$, which equals to $\| \x_a - \x_p \| / 2$.

Inspired by the triplet loss (Eq.~\ref{eq:triplet_loss}), we seek for the optimum embedding such that the samples of different classes can be separated well as the angular constraint (Eq.~\ref{eq:angular_constraint2}) describes.
In a nutshell, our angular loss consists of minimizing the following hinge loss,
\begin{align}
  l_{ang} (\cT) = \Big[ \|\x_a - \x_p \|^{2} - 4 \tan^{2} \alpha \|\x_n - \x_c \|^{2} \Big]_+. \label{eq:angular_loss}
\end{align}
To better understand the effect of optimizing the angular loss, we can investigate the gradient of $l_{ang}$ with respect to $\x_a$, $\x_p$ and $\x_n$, which are
\begin{align} \label{eq:angular_gradient}
& \frac{\partial l_{ang}(\cT)}{\partial \x_a} = 2 (\x_a-\x_p) - 2 \tan^{2} \alpha (\x_a+\x_p-2\x_n), \nonumber \\
& \frac{\partial l_{ang}(\cT)}{\partial \x_p} = 2 (\x_p-\x_a) - 2 \tan^{2} \alpha (\x_a+\x_p-2\x_n), \nonumber \\
& \frac{\partial l_{ang}(\cT)}{\partial \x_n} = 4 \tan^{2} \alpha \Big[ (\x_a + \x_p) - 2\x_n \Big],
\end{align}
if $\angle n'$ is larger than $\alpha$, or zero otherwise.
As illustrated in Fig.~\ref{fig:toy_example}b, the gradient pushes the negative point $\x_n$ away from $\x_c$, the center of local cluster defined by $\x_a$ and $\x_p$.
In addition, the anchor $\x_a$ and the positive $\x_p$ are dragged towards each other.
Compared to the original triplet loss whose gradients (Eq.~\ref{eq:triplet_gradient}) only depend on two points, the gradients in Eq.~\ref{eq:angular_gradient} are much more robust as they consider all the three points simultaneously.





\subsection{Implementation details}

\begin{figure}
\centering\includegraphics[width=.4\textwidth]{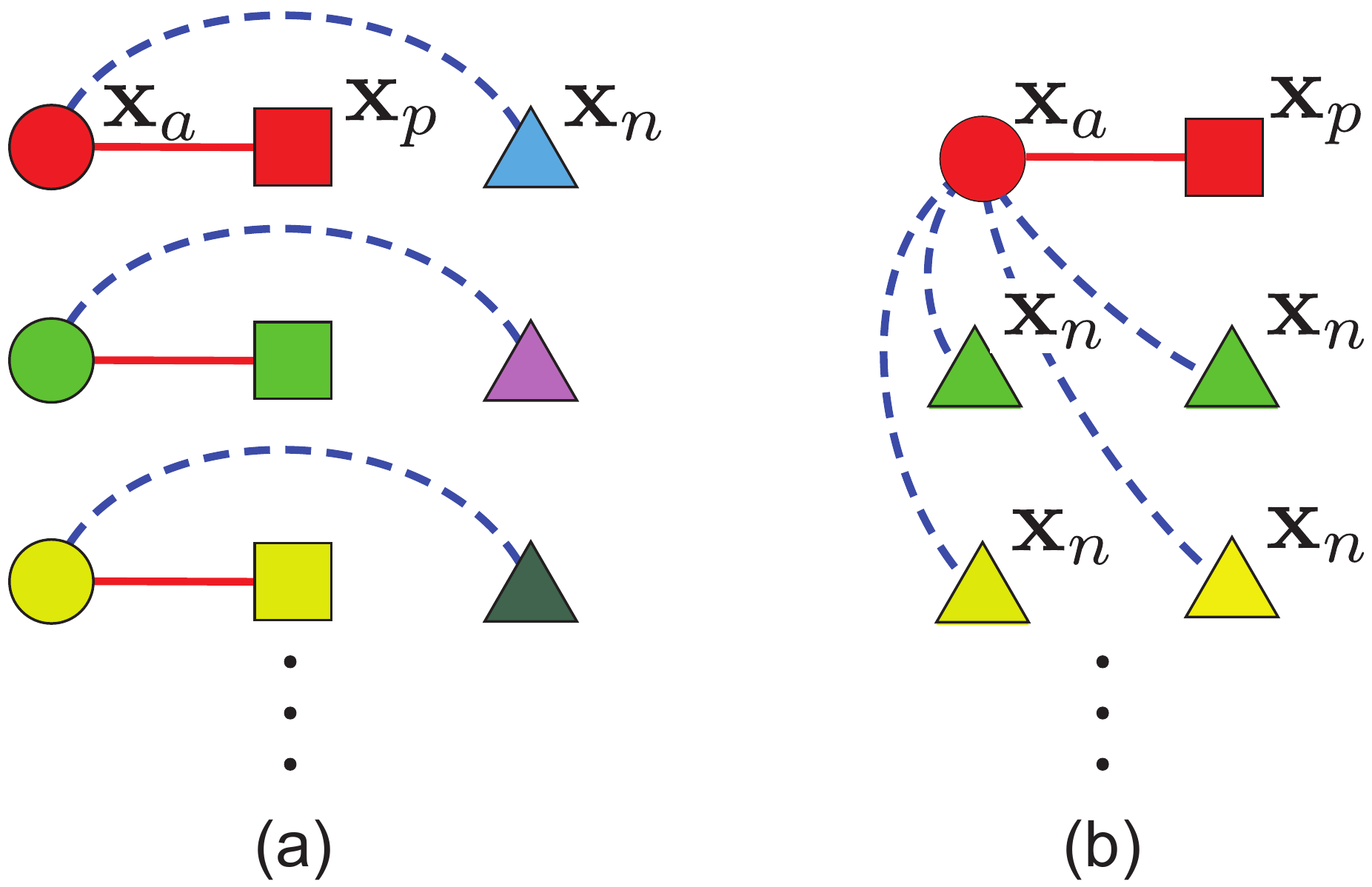}
\caption{Comparison between different sampling methods. For each node, we use color to indicate the class label and shape for its role (\ie, anchor, positive or negative) in triplet. (a) Traditional triplet sampling. (b) N-pair sampling. To keep plot clean, we only show the connection inside one tuplet.} \label{fig:sampling}
\end{figure}

Eq.~\ref{eq:angular_loss} defines the angular loss on a triplet.
When optimizing a mini-batch containing multiple triplets, we found our method can be further improved in two ways.

First, we enhance the mini-batch optimization by making the full use of the batch.
As illustrated in Fig.~\ref{fig:sampling}a, the conventional sample strategy constructs a mini-batch as multiple disjoint triplets without interaction among them.
This poses a large bottleneck in optimization as it can only encode a limited amount of information.
To allow joint comparison among all samples in the batch, we follow the sampling strategy proposed in N-pair loss~\cite{Sohn16} to construct tuplets with multiple negative points.
More concretely, we first draw $N / 2$ different classes, from each of which we then randomly sample two training images.
The main benefit behind N-pair sampling is that it can avoid the quadratic possible combinations of tuplets.
For instance, as shown in Fig.~\ref{fig:sampling}b, given a batch with $N$ samples $\cB = \{\x_i, y_i\}_{i=1}^N$, there are in total $N$ tuplets, each of which is composed by a pair of anchor $\x_a \in \cB$ and positive $\x_p \in \cB$ of the same class, and $N - 2$ negative from other classes.

Second, a direct extension of Eq.~\ref{eq:angular_loss} to consider more than one negative point would result in a very non-smooth objective function.
Inspired by recent work~\cite{SongXJS16,Sohn16,SongJR016}, we replace the original hinge loss with its smooth upper bound, \ie, $\log(\exp(y_1) + \exp(y_2)) \geq \max(y_1, y_2)$.
By assuming feature is of unit length (\ie, $\| \x \|=1$) in Eq.~\ref{eq:angular_loss}, we derive the angular loss for a batch $\cB$ using the following \emph{log-sum-exp} formulation:
\begin{align}
  l_{ang}(\cB) &= \frac{1}{N} \sum_{\x_a \in \cB}\bigg\{\log\bigg[1
                 +\sum_{\substack{\x_n \in \cB \\ y_n \neq y_a, y_p}}\exp \big(f_{a,p,n}\big)\bigg] \bigg\}, \label{eq:angular_loss_batch}
\end{align}
where in $f_{a,p,n}$, we drop the constant terms depending on the value of $\|\x\|$ in a similar spirit to N-pair loss~\cite{Sohn16}, \ie,
\begin{align}
  f_{a,p,n} = 4\tan^{2}\alpha(\x_a + \x_p)^T \x_n - 2(1 + \tan^{2}\alpha) \x_a^T \x_p. \nonumber
\end{align}




Our work on angular loss explores the third-order relations beyond the scope of the well-studied pair-wise distance.
Due to its flexibility and generality, we can easily combine the angular constraint with traditional distance metric loss to boost the overall performance.
As an example, we mainly investigate the combination with the N-pair loss~\cite{Sohn16}, one of the latest work for deep metric learning,
\begin{align} \label{eq:npair_angular_loss}
l_{npair \& ang}(\cB) = l_{npair}(\cB) + \lambda l_{ang}(\cB),
\end{align}
where $l_{npair}(\cB)$ denotes the original N-pair loss as,
\begin{align}
  l_{npair}(\cB) = & \frac{1}{N} \sum_{\x_a \in \cB}\bigg\{ \log\bigg[ 1 + \nonumber \\
  & \quad \sum_{\substack{\x_n \in \cB\\ y_n \neq y_a, y_p}}\exp\big(\x_a^T \x_n -\x_a^T \x_p \big)\bigg] \bigg\}, \label{eq:npair_loss}
\end{align}
and $\lambda$ is a trade-off weight between N-pair and the angular loss.
In all experiments, we always set $\lambda = 2$ as it consistently yields promising result.

\section{Experiments}

In this section, we evaluate deep metric learning algorithms on both image retrieval and clustering tasks.
Our method has been shown to achieve state-of-the-art performance on three public benchmark datasets.

\subsection{Benchmark datasets}

We conduct our experiments on three public benchmark datasets. For all datasets, we follow the conventional protocol of splitting training and testing:

\textbf{CUB-200-2011}~\cite{BransonHWPB14} dataset has 200 species of birds with 11,788 images included, where the first 100 species (5,864 images) are used for training and the remaining 100 species (5,924 images) are used for testing.

\textbf{Stanford Car}~\cite{KrauseSDF013} dataset is composed by 16,185 cars images of 196 classes. We use the first 98 classes (8,054 images) for training and the other 98 classes (8,131 images) for testing.

\textbf{Online Products}~\cite{SongXJS16} dataset contains 22,634 classes with 120,053 product images in total, where the first 11,318 classes (59,551 images) are used for training and the rest classes (60,502 images) are used for testing.

\subsection{Baselines}

In order to evaluate the superiority of the proposed method, we compare with three baselines:

\textbf{Triplet Loss}: We implement the standard triplet embedding by optimizing Eq.~\ref{eq:triplet_loss}.
To be fair in comparison, we apply triplet loss embedding with two sampling strategies.
Following the most standard setting, the mini-batch of \textbf{Triplet-I (T-I)} was constructed by sampling disjoint triplets as illustrated in Fig.~\ref{fig:sampling}a.
In the second case of \textbf{Triplet-II (T-II)}, we optimize Eq.~\ref{eq:triplet_loss} using the N-pair sampling as shown in Fig.~\ref{fig:sampling}b to keep consistent with the angular loss.

\textbf{Lifted Structure (LS)}~\cite{SongXJS16}: We adopt the open-source code from the authors' website with the default parameters used in the paper.

\textbf{N-pair Loss (NL)}~\cite{Sohn16}: We implement N-pair loss (Eq.~\ref{eq:npair_loss}) closely following the illustration of the paper. We found our implementation achieved similar results as reported in the paper.

For our method, we implement two versions, \textbf{Angular Loss (AL)} and \textbf{N-pair \& Angular Loss (NL\&AL)}, that optimize Eq.~\ref{eq:angular_loss_batch} and Eq.~\ref{eq:npair_angular_loss} respectively. To be comparable with prior work, we employ the N-pair sampling (Fig.~\ref{fig:sampling}b) shared by the baselines of \textbf{Triplet-II} and \textbf{N-pair Loss}.

As the focus of this work is the similarity measure, we did not employ any hard negative mining strategies to complicate the comparison.
But it is worth mentioning that our work can be easily combined with any hard negative mining method.

\subsection{Evaluation metrics}

Following the standard protocol used in~\cite{SongXJS16,Sohn16}, we evaluate the performance of different methods in both retrieval and clustering tasks.
We split each dataset into two sets of disjoint classes,
one for training and the other for testing the retrieval and clustering performance of the unseen classes.
For retrieval task, we calculate the percentage of the testing examples whose $R$ nearest neighbors contain at least one example of the same class.
This quantity is also known as Recall@R, the defacto metric~\cite{JegouDS11} for image retrieving evaluation.
For clustering evaluation, we adopt the code from~\cite{SongXJS16} by clustering testing examples using the k-means algorithm.
The quality of clustering is reported in terms of the standard F$_1$ and NMI metrics. See~\cite{SongXJS16} for their detailed definition.

\subsection{Training setup}

The Caffe package~\cite{JiaSDKLGGD14} is used throughout the experiments.
All images are normalized to 256-by-256 before further processing.
The embedding size is set to $D = 512$ for all embedding vectors, and no normalization is conducted before computing loss.
We omit the comparison on different embedding sizes as the performance change is minor.
This fact is also evidenced in~\cite{SongXJS16}.
GoogLeNet~\cite{SzegedyLJSRAEVR15} pretrained on ImageNet ILSVRC dataset~\cite{RussakovskyDSKS15} is used for initialization and a randomly initialized fully connected layer is added.
The new layer is optimized with 10 times larger learning rate than the other layers.
We fix the base learning rate to $10^{-4}$ for all datasets except for the CUB-200-2011 dataset, for which we use a smaller rate $10^{-5}$ as it has fewer images and is more likely to meet the overfitting problem.
We use SGD with 20k training iterations and 128 mini-batch size.
Standard random crop and random horizontal mirroring are used for data augmentation.
Notice that our method incurs negligible computational cost compared to traditional triplet loss. Therefore, the training time is almost same as other baselines.

\subsection{Result analysis}

Tables~\ref{tab:cub}, \ref{tab:car} and \ref{tab:product} compare our method with all baselines in both clustering and retrieval tasks.
These tables show that the two recent baselines, lifted structure (LS)~\cite{SongXJS16} and N-pair loss (NL)~\cite{Sohn16}, can always improve the standard triplet loss (T-I and T-II).
In particular, N-pair achieves a larger margin in improvement because of the advance in its loss design and batch construction.
Compared to previous work, the proposed angular loss (AL) consistently achieves better results on all three benchmark datasets.
It is important to notice that the proposed angular loss (AL) employs the same sampling strategies as triplet loss (T-II) and N-pair loss (NL).
This clearly indicates the superiority of the new loss for solving deep metric learning problem.
By integrating with the original N-pair loss, the joint optimization of angular loss in NL\&AL can lead to the best performance among all the methods in all metrics.

\begin{figure*}
\centering\includegraphics[width=\textwidth]{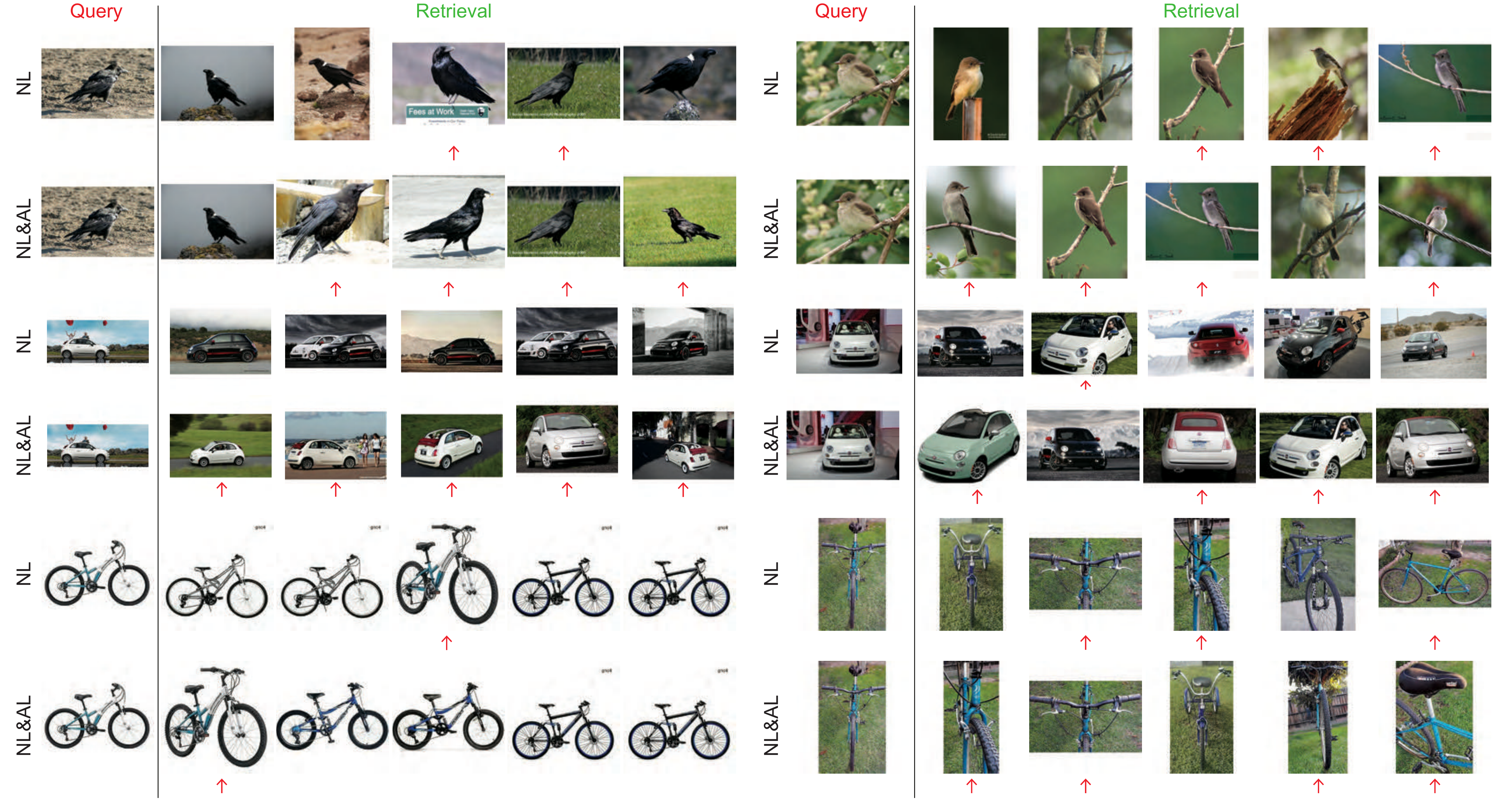}
\caption{Comparison of queries and top-5 retrievals between N-pair (NP) and our method (NP\&AL). From top to bottom, we plot two examples for the CUB-200-2011, Stanford car and online products dataset respectively. The retrieved images pointed by an arrow are the ones that belong to the same class as the query.} \label{fig:query}
\end{figure*}

\begin{figure*}
\centering\includegraphics[width=\textwidth]{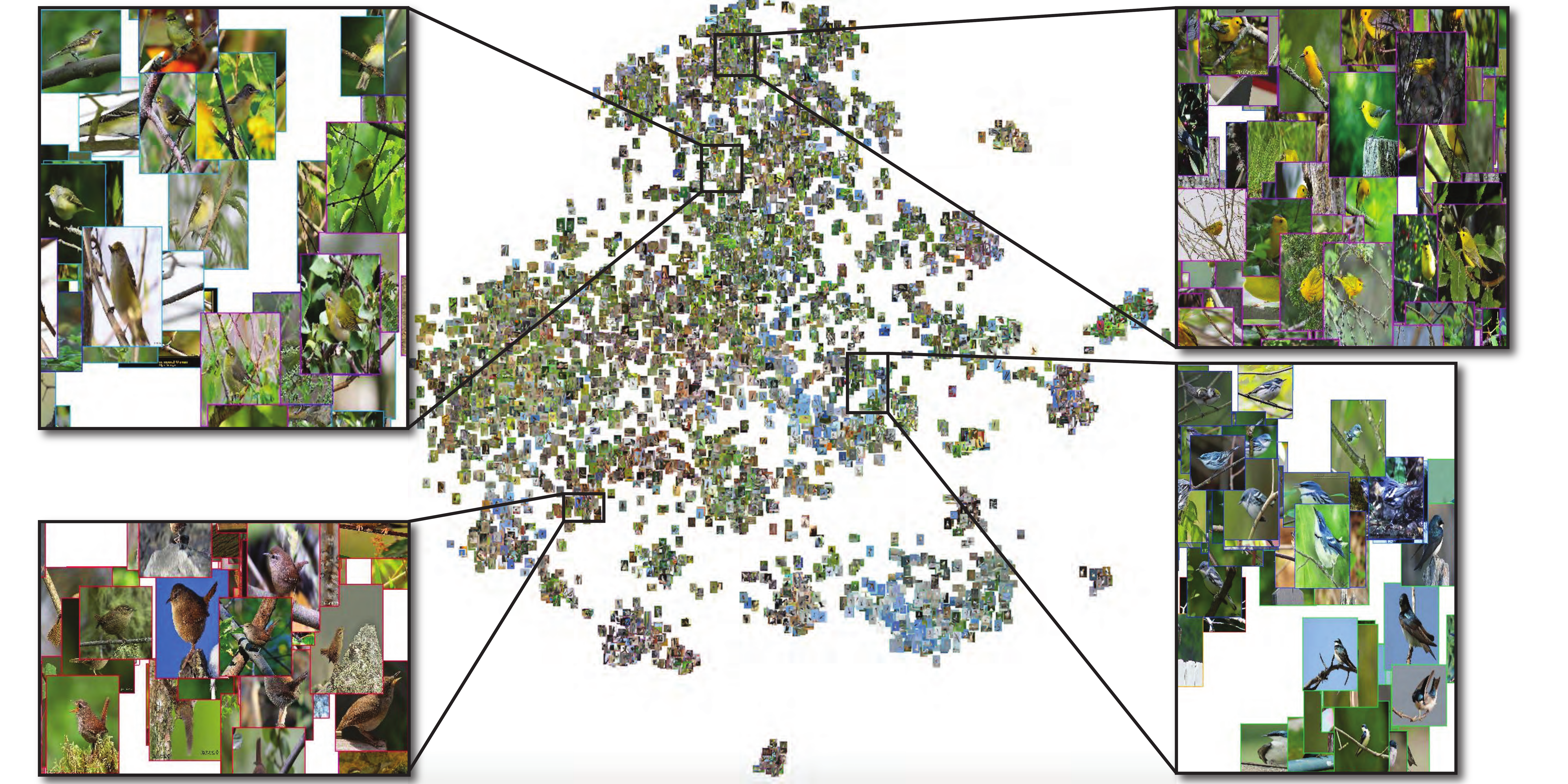}
\caption{Visualization of feature embedding computed by our method (NP\&AL) using t-SNE on the CUB-200-2011 dataset.} \label{fig:tsne}
\end{figure*}

Fig.~\ref{fig:query} compares NL\&AL with N-pair loss on the task of image retrieval.
As it can be observed, the proposed NL\&AL learns a more discriminative feature that helps in identifying the correct images especially when the intra-class variance is large.
For example, given a query image of FIAT 500 Convertible 2012 at the fourth row of Fig.~\ref{fig:query} on the right side, the top-5 images retrieved by NL\&AL contain four successful matches that belong to the same class as the query, while N-pair method fails to identify them.
In addition, Fig.~\ref{fig:tsne} visualizes the feature embedding computed by our method (NL\&AL) in 2-D using t-SNE~\cite{Maaten14}.
We highlight several representative classes by enlarging the corresponding regions in the corners.
Despite the large pose and appearance variation, our method effectively generates a compact feature mapping that preserves semantic similarity.

A key parameter of our method is the margin $\alpha$, that determines to what degree the constraint (Eq.~\ref{eq:angular_constraint2}) would be activated.
Table~\ref{tab:alpha} and Table~\ref{tab:alpha2} study the impact of choosing different $\alpha$ for the retrieval task on the Stanford car and online product datasets, respectively.
Choosing $\alpha=45^{\circ}$ for Stanford car and $\alpha=36^{\circ}$ for online product lead to the best performance for the method of NL\&AL.
We found that our method performs consistently well in all three dataset for $36^{\circ} \leq \alpha \leq 55^{\circ}$.
It deserves to be mentioned that, without integrating with NL, AL preforms comparably with NL, and even better when mining a proper value of $\alpha$, which is shown in Table~\ref{tab:alpha2}.


\begin{table}
\begin{center}
\begin{threeparttable}
\begin{tabular}{p{1cm}ccccccc}
\toprule
\multirow{2}{*}{Method}&
\multicolumn{2}{c}{Clustering (\%)}&
\multicolumn{4}{c}{Recall@R (\%)}\cr
\cmidrule(lr){2-3}\cmidrule(lr){4-7}
&NMI&F$_1$&R=1&R=2&R=4&R=8\cr
\midrule
T-I & $53.7$ & $19.7$ & $42.2$ & $54.4$ & $66.2$ & $76.7$ \cr
T-II & $54.1$ & $20.0$ & $42.8$ & $54.9$ & $66.2$ & $77.6$ \cr
LS & $56.2$ & $22.7$ & $46.5$ & $58.1$ & $69.8$ & $80.2$ \cr
NL & $60.2$ & $28.2$ & $51.9$ & $64.3$ & $74.9$ & $83.2$ \cr
AL & $61.0$ & $\mathbf{30.2}$ & $53.6$ & $65.0$ & $75.3$ & $83.7$ \cr
NL\&AL & $\mathbf{61.1}$ & $29.4$ & $\mathbf{54.7}$ & $\mathbf{66.3}$ & $\mathbf{76.0}$ & $\mathbf{83.9}$ \cr
\bottomrule
\end{tabular}
\end{threeparttable}
\newline
\caption{Comparison of clustering and retrieval on the CUB-200-2011~\cite{BransonHWPB14} dataset.} \label{tab:cub}
\end{center}
\end{table}

\begin{table}
\begin{center}
\begin{threeparttable}
\begin{tabular}{p{1cm}ccccccc}
\toprule
\multirow{2}{*}{Method}&
\multicolumn{2}{c}{Clustering (\%)}&
\multicolumn{4}{c}{Recall@R (\%)}\cr
\cmidrule(lr){2-3}\cmidrule(lr){4-7}
&NMI&F$_1$&R=1&R=2&R=4&R=8\cr
\midrule
T-I & $53.8$ & $18.7$ & $45.5$ & $59.0$ & $71.0$ & $80.8$ & $$ \cr
T-II & $54.3$ & $19.6$ & $46.3$ & $59.9$ & $71.4$ & $81.3$ \cr
LS & $55.1$ & $21.5$ & $48.3$ & $61.1$ & $71.8$ & $81.1$ \cr
NL & $62.7$ & $31.8$ & $68.9$ & $78.9$ & $85.8$ & $90.9$ \cr
AL & $62.4$ & $31.8$ & $71.3$ & $80.7$ & $87.0$ & $91.8$ \cr
NL\&AL & $\mathbf{63.2}$ & $\mathbf{32.2}$ & $\mathbf{71.4}$ & $\mathbf{81.4}$ & $\mathbf{87.5}$ & $\mathbf{92.1}$ \cr
\bottomrule
\end{tabular}
\end{threeparttable}
\newline
\caption{Comparison of clustering and retrieval on the Stanford car~\cite{KrauseSDF013} dataset.} \label{tab:car}
\end{center}
\end{table}

\setlength{\tabcolsep}{0.5em}
\begin{table}
\begin{center}
\begin{threeparttable}
\begin{tabular}{p{1cm}ccccccc}
\toprule
\multirow{2}{*}{Method}&
\multicolumn{2}{c}{Clustering (\%)}&
\multicolumn{4}{c}{Recall@R (\%)}\cr
\cmidrule(lr){2-3}\cmidrule(lr){4-7}
&NMI&F$_1$&R=1&R=10&R=100&R=1000\cr
\midrule
T-I & $86.2$ & $19.9$ & $56.5$ & $74.7$ & $88.3$ & $96.2$\cr
T-II & $86.4$ & $21.0$ & $58.1$ & $76.0$ & $89.1$ & $96.4$\cr
LS & $87.4$ & $24.7$ & $63.0$ & $80.5$ & $91.7$ & $97.5$\cr
NL & $87.7$ & $26.3$ & $66.9$ & $83.0$ & $92.3$ & $97.7$\cr
AL & $87.8$ & $26.5$ & $67.9$ & $83.2$ & $92.2$ & $97.7$\cr
NL\&AL & $\mathbf{88.6}$ & $\mathbf{29.9}$ & $\mathbf{70.9}$ & $\mathbf{85.0}$ & $\mathbf{93.5}$ & $\mathbf{98.0}$\cr
\bottomrule
\end{tabular}
\end{threeparttable}
\newline
\caption{Comparison of clustering and retrieval on the online products~\cite{SongXJS16} dataset.} \label{tab:product}
\end{center}
\end{table}

\begin{table}
\begin{center}
\begin{threeparttable}
\begin{tabular}{ccccc}
\toprule
\multirow{2}{*}{NL\&AL($\alpha$)}&
\multicolumn{4}{c}{Recall@R (\%)}\cr
\cmidrule(lr){2-5}
&R=1&R=2&R=4&R=8\cr
\midrule
$\alpha=36^{\circ}$ & $69.9$ & $79.7$ & $86.8$ & $91.8$\cr
$\alpha=42^{\circ}$ & $70.7$ & $80.5$ & $87.2$ & $91.9$\cr
$\alpha=45^{\circ}$ & $\mathbf{71.4}$ & $\mathbf{81.4}$ & $\mathbf{87.5}$ & $\mathbf{92.1}$\cr
$\alpha=48^{\circ}$ & $71.3$ & $80.4$ & $87.0$ & $91.9$\cr
$\alpha=55^{\circ}$ & $69.0$ & $78.1$ & $85.3$ & $90.8$\cr
\bottomrule
\end{tabular}
\end{threeparttable}
\newline
\caption{Comparison of different values on $\alpha$ for our method on Stanford car dataset.} \label{tab:alpha}
\end{center}
\end{table}

\begin{table}
\begin{center}
\begin{threeparttable}
\begin{tabular}{cccc}
\toprule
Method & NL & NL\&AL($\alpha=45^{\circ}$) & NL\&AL($\alpha=36^{\circ}$) \cr
Recall@1 (\%) & $66.9$ & $69.2$ & $\mathbf{70.9}$ \cr
\midrule
Method & NL & AL($\alpha=45^{\circ}$) & AL($\alpha=36^{\circ}$) \cr
Recall@1 (\%) & $66.9$ & $66.4$ & $\mathbf{67.9}$ \cr
\bottomrule
\end{tabular}
\end{threeparttable}
\newline
\caption{Comparison of different values on $\alpha$ for our method on the online product dataset.} \label{tab:alpha2}
\end{center}
\end{table}

\section{Conclusion}

In this paper, we propose a novel angular loss for deep metric learning.
Unlike most methods that formulate objective based on distance, we resort to constrain the angle of the triplet triangle in the loss.
Compared to pair-wise distance, angle is a rotation and scale invariant metric, rendering the objective more robust against the large variation of feature map in real data.
In addition, the value of angle encodes the triangular geometry of three points simultaneously.
Given the same triplet, it offers additional source of constraints to ensure that dis-similar points can be separated.
Furthermore, we show how the angular loss can be easily integrated into other frameworks such as N-pair loss~\cite{Sohn16}.
The superiority of our method over existing state-of-the-art work is verified on several benchmark datasets.

In the future, we hope to extend our work in two directions.
First, our method origins from the triplet loss and leverages the third-order relation among three points.
It is interesting to consider more general case with four or more samples.
Previous work~\cite{ZhangZLZ16,HuangLT16} studied the case of quadruplet but still employed certain distance-based objectives.
One possible extension of our idea on quadruplet is to construct a triangular pyramid and constrain the angle between the side edge and the plane on the bottom.
Second, it is beneficial to combine our method with other practical tricks such as hard negative mining~\cite{YuanYZ16} or new clustering-like frameworks~\cite{ref7_lx,SongJR016}.

{\small
\bibliographystyle{ieee}
\bibliography{egbib}
}

\end{document}